\documentclass[10pt, a4paper]{article}
\usepackage[affil-it]{authblk} 
\usepackage{etoolbox}
\usepackage{lmodern}
\usepackage{url}
\usepackage{times}
\usepackage{latexsym}
\usepackage{tikz}
\usetikzlibrary{shapes,arrows,chains}
\usepackage{tikz-dependency}
\usepackage{url}
\usepackage{multirow}
\usepackage{array}
\usepackage{subcaption}
\usepackage{float}
\usepackage{setspace}
\usepackage{verbatim}
\usepackage{lingmacros}
\usepackage{forest}
\useforestlibrary{linguistics}
\forestapplylibrarydefaults{linguistics}
\usepackage{linguex}
\usepackage{booktabs}
\usepackage{CJKutf8}
\usepackage{times}
\usepackage{geometry}
\usepackage{sectsty}
\usepackage{natbib}
\forestset{
sm edges/.style={for tree={parent anchor=south, child anchor=north,base=bottom},
                 where n children=1{tier=word}{}
                 },
}   

\forestset{default preamble'={
    for tree={align=center,parent anchor=south, child anchor=north,base=bottom},
    before drawing tree={
      sort by=y,
      for min={tree}{baseline}
    }
}}

\makeatletter
\patchcmd{\@maketitle}{\LARGE \@title}{\fontsize{14}{19.2}\selectfont\@title}{}{}
\makeatother

\title{Is Argument Structure of Learner Chinese Understandable: \\A Corpus-Based Analysis} 
\author{Yuguang Duan$^\clubsuit$,  Zi Lin$^{\spadesuit\clubsuit}$ and Weiwei Sun$^{\spadesuit\heartsuit}$
	}
\affil{$^\spadesuit$Institute of Computer Science and Technology, Peking University \\
	$^\spadesuit$The MOE Key Laboratory of Computational Linguistics, Peking University \\
    $^\clubsuit$Department of Chinese Literature and Linguistics\\
        $^\heartsuit$Center for Chinese Linguistics, Peking University\\
    \protect\url{{ariaduan,zi.lin,ws}@pku.edu.cn}\\
    {\bf Bio data}: Our lab focuses on natural language processing and language engineering(NLP). \\
    We develop computational resources, including semantic and syntactic parsers and annotated corpora.}
\date{}
\sectionfont{\fontsize{10}{15}\selectfont}
\subsectionfont{\fontsize{10}{15}\selectfont}

\begin{document}

\maketitle

\begin{abstract}
This paper presents a corpus-based analysis of argument structure errors in learner Chinese.
The data for analysis includes sentences produced by language learners as well as their corrections by native speakers.
We couple the data with semantic role labeling annotations that are manually created by two senior students whose majors are both Applied Linguistics.
The annotation procedure is guided by the Chinese PropBank specification, which is originally developed to cover first language phenomena. 
Nevertheless, we find that it is quite comprehensive for handling second language phenomena.
The inter-annotator agreement is rather high, suggesting the understandability of learner texts to native speakers. 
Based on our annotations, we present a preliminary analysis of competence errors related to argument structure.
In particular, speech errors related to word order, word selection, lack of proposition, and argument-adjunct confounding are discussed.

\noindent {\bf Keywords}: Learner Chinese, competence error, argument structure, semantic role labeling

\noindent {\bf Highlights}: \shortex{1}
{$\bullet$ We build an L2-L1 parallel corpus with semantic role labeling.}
{$\bullet$ We apply this corpus to analysis on learner texts.}
{$\bullet$ Five typical errors related to argument structure are discussed.}
\\
 \vspace{6pt}


\end{abstract}

\begin{CJK*}{UTF8}{gbsn}
\section{Introduction}
\label{sec:intro}
Corpus-based linguistic analysis is one of the fastest-growing methodologies in contemporary linguistics. 
It has been applied to studying not only the patterns of native speakers' language (henceforth, ``L1''), but also the typical competence errors of foreign learners' language (henceforth, ``L2''). 

A learner corpus collects the language produced by people learning a foreign language, which have been essential for building natural language Processing (NLP) systems related to learner languages, as reported by \citep{nagata-sakaguchi:2016:P16-1,berzak-EtAl:2016:P16-1}. Furthermore, L2-L1 parallel treebanks have been shown beneficial for learner language analysis \citep{l1l2treebank}.

Presently, most learner corpora aim to provide {\it grammatical error} labeling as well as correction. This can be helpful to second language teaching to a certain degree, but not comprehensive enough for more intelligent tasks, such as automatic essay scoring and determining the native language of a language learner. 
In such tasks, we not only need to spot the grammatical errors, but also need to scrutinize the logic and semantic propriety to evaluate the coherency of the discourse structure, the persuasiveness of the argumentation, etc. 
This requires a corpus with more exquisite syntactic and semantic labels, e.g., argument structure, that supports various Machine Learning algorithms for building NLP systems. 
This will also facilitate the probe to languages: statistical models can be employed to analyze learner corpora to provide insights into the nature of language acquisition or typical learner needs. 

In this paper, we build an L2-L1 parallel corpus which has scrutinized semantic role labeling (SRL), and apply it to analysis on learner texts. 
We present a preliminary analysis of competence errors related to argument structure.
In particular, speech errors related to word order, word selection, lack of proposition, and argument-adjunct confounding are discussed.

\section{Background}


SRL is a widely-studied NLP task that assigns semantic role labels to words or phrases in a sentence that indicate the argument structures.
It consists of detecting the semantic arguments associated with the predicate of a sentence and assigning semantic roles to them according to their relationship to that predicate. Typical semantic roles include {\it Agent}, {\it Patient}, {\it Source}, {\it Goal}, and so forth,
which are core arguments to a predicate, as well as {\it Location}, {\it Time}, {\it Manner}, {\it Cause}, and so on, which are adjuncts.

SRL is important to understand the essential meaning of the original input language sentences -- {\it who} did {\it what} to {\it whom}, for {\it whom} or {\it what}, {\it how}, {\it where}, {\it when} and {\it why}, for it provides sentence-level semantic analysis of text that characterizes events.

The Chinese PropBank \citep[CPB;][]{ChPropBank} is a popular semantically annotated corpus for research on Chinese SRL.
It adds a layer of predicate--argument structures to the Chinese TreeBank, assigning semantic role labels to syntactic constituents (rather than to the headwords in a dependency structure) in a sentence.
Each verb has several {\it framesets} that are annotated with a fixed number of arguments: 
the core arguments of a predicate are labeled
with a contiguous sequence of integers, in the form of \textit{AN} (\textit{N} is a natural number); the adjuncts are annotated with the label \textit{AM} followed by a secondary tag that provides semantic information such as location, manner, and time. All the labels are defined by a general set of guidelines.

\section{Methodology of the Study}
\label{sec:human}
Motivated by the importance of corpus in both (quantitative) linguistic analysis and building NLP systems, we are concerned with constructing a semantic role-annotated L2 corpus.
To this end, we need to gather a L2 corpus in advance. 
In this paper, we use Lang-8, which contains large-scale learner texts of Mandarin Chinese that are collected from ``language exchange" social networking services (SNS), a language-learning website where native speakers freely choose learners' essays to correct. 
The collecting work was done by our lab member Yuanyuan Zhao \citep{nlpcc2018gec}, following \citep{mizumoto:2011}.

To make sure that a L2 corpus with semantic role labels is achievable, we first examined whether the learner texts can be understood by native speakers. To this end, we first conducted an inter-annotation between two annotators whose majors are Applied Linguistics to see if a high agreement can be achieved.
In this process, we created a corpus consisting of manually-annotated predicate--argument labels on 600 L2-L1 pairs for learner Chinese. 

We also notice that mother languages of language learners have a great impact on grammatical errors and hence influence the following ontological study and automatic semantic analysis. Therefore, our corpus includes four typologically different languages, i.e., English (ENG), Japanese (JPN), Russian (RUS) and Arabic (ARA). Each has a sub-corpus consisting of 150 sentence pairs.

Our annotators first annotated 50 parallel sentences for each native language, adapting PropBank specification as annotation heuristics, and then produced an initial adjudicated gold standard according to these 400 sentences. Based on this gold standard, the annotators proceeded to annotate a 100-sentence set for each language. The inter-annotator agreement is reported on these larger sets.
\section{Findings}
\label{sec:human}



\begin{table}[H]
\centering
\caption{Inter-annotator agreement.}
\scalebox{1.0}{
\begin{tabular}{llccc}
\hline
        &&     P    &     R   &     F   \\\hline
\multirow{2}{*}{ENG}&  L1     &   95.87 &  96.17 &  96.02\\
&L2     &   94.78 &  93.06 &  93.91\\\hline
\multirow{2}{*}{JPN} &L1     &   97.95 &  98.69 &  98.32\\
&L2     &   96.07 &  97.48 &  96.77\\\hline
\multirow{2}{*}{RUS}&L1     &   96.95 &  95.41 &  96.17\\
 & L2     &   97.04 &  94.08 &  95.53\\\hline
\multirow{2}{*}{ARA}&L1     &   96.95 &  97.76 &  97.35\\
 & L2     &   97.12 &  97.56 &  97.34\\
\hline                       
\end{tabular}}
\label{tb:agreement}
\end{table}

We calculate the precision (P), recall (R), F-score (F) to measure the inter-annotator agreement, as shown in Table \ref{tb:agreement}.
The inter-annotator agreement indicates that semantic annotations between the two annotators for both L1 and L2 sentences are quite consistent. 
All L1 texts have F-scores above 95, comparable to the annotation of CPB \citep{ChPropBank}. 
We take this result as a reflection that our annotators are qualified. 
F-scores of L2 sentences are all above 90, just a little bit lower than those of L1, indicating that L2 sentences can be greatly understood by native speakers.

\begin{table}[H]
\centering
\caption{Inter-annotator agreement (F-scores) relative to languages and role types.}
\scalebox{1.0}{
\begin{tabular}{llcccc}
\hline
                    &    & ENG    & JPN    & RUS    & ARA    \\ \hline
\multirow{6}{*}{L1} & \textit{A0} & 97.23  & 99.10  & 97.66  & 98.22  \\
                    & \textit{A1} & 96.70  & 96.99  & 98.05  & 98.34  \\
                    & \textit{A2} & 88.89  & 100.00 & 100.00 & 92.59  \\
                    & \textit{A3} & 100.00 & 100.00 & 100.00 & 100.00 \\
                    & \textit{A4} & 100.00 & -      & -      & 100.00 \\
                    & \textit{AM} & 94.94  & 98.35  & 93.07  & 96.02  \\ \hline
\multirow{6}{*}{L2} & \textit{A0} & 94.09  & 95.77  & 97.92  & 97.88  \\
                    & \textit{A1} & 90.68  & 97.93  & 97.40  & 98.68  \\
                    & \textit{A2} & 88.46  & 100.00 & 95.24  & 93.33  \\
                    & \textit{A3} & 100.00 & 100.00 & 100.00 & -      \\
                    & \textit{A4} & 100.00 & -      & -      & -      \\
                    & \textit{AM} & 96.97  & 96.51  & 91.78  & 96.02  \\ 
\hline
\end{tabular}}
\label{tb:agreement-detail}
\end{table}

Table \ref{tb:agreement-detail} further reports agreements on each argument (\textit{AN}) and adjunct (\textit{AM}) in detail, according to which the high scores are attributed to the high agreement on arguments (\textit{AN}). 
The labels of \textit{A3} and \textit{A4} have no disagreement since they are sparse in CPB and are usually used to label specific semantic roles that have little ambiguity.

\section{Discussion and Implications}
\subsection{Disagreement on \textit{A2} in Learner English}
From Table \ref{tb:agreement} and \ref{tb:agreement-detail}, we notice that the F-score of English L2 (93.91) is relatively low compared to the other L2s for the low agreement on \textit{A2} (88.46). 
We find that most L2 sentences with \textit{A2} disagreements appear to have a mismatch between the Chinese and English attributive clause syntax. Take sentence in (1a) for an example.

\eenumsentence{
\footnotesize
\item \shortex{10}
     {我 & 帮 & 他们 & 盖 & 了 & 一个 & 盒子 & 可以 & 装满 & 泥土。} 
     {I & help.\textsc{past} & they.\textsc{acc} & make & \textsc{asp} & a & box & can & fill & soil. } 
     {I helped them make a box that can be filled with soil.}
\item \shortex{11}
     {我 & 帮 & 他们 & 盖 & 了 & 一个 & 可以 & 装满 & 泥土 & 的 & 盒子。 } 
     {I & help.\textsc{past} & they.\textsc{acc} & make & \textsc{asp} & a & can & fill & soil & DE & box. } 
     {I helped them make a box that can be filled with soil.}
     
\item \shortex{4}
     {{[我]}$_{A0}$ & {[帮]}$_{rel}$ & {[他们]}$_{A1}$ & {[盖了一个盒子可以装满泥土]}$_{A2}$。 } 
     {{[我]}$_{A0}$ & {[帮]}$_{rel}$ & {[他们]}$_{A1}$ & {[盖了一个盒子]}$_{A2}$可以装满泥土。 } 
     {}
}

In the sentence, 装满泥土的盒子 (``a box that can be filled with soil'') should be treated as a whole and labeled with \textit{A2} (thing \textit{A0} helps \textit{A1} with), as 装满泥土（``that can be filled with soil''） is the attributive clause of 盒子 （``a box''） in English grammar.
However, in Chinese, it should be written like (1b) where the attributive elements are put in front of the noun appended with an auxiliary word 的. 
This syntactic difference between the two languages causes the annotators to have splitting ideas on the boundaries of \textit{A2}, as shown in (1c).

\subsection{Disagreement on \textit{AM} in all L2 Sentences}
Another source of disagreement is the labels of \textit{AM}. We analyze those L2 sentences with different annotated labels from two annotators, and find five predominant types of error, as described in Table \ref{tb:AM error types}. 

\paragraph{Word order} 
This error occurs when the sentence switches the position of constituents.
In the example, 离开鄂木斯克 (``leave Emusike'') is the object of 打算 (``try to''), so 别 (``don't'') should be labeled as a negative adjunct of 打算 (``try to''), 
while in the learner sentence 离开鄂木斯克 (``leave Emusike'') is transited forward, causing 别 (``don't') to become the adjunct of 离开 (``leave'') according to the principle of proximity. 
The unconformity between semantic relationship and syntactic structure can easily lead to disagreement, as shown in Table \ref{tb:AM error types}. This type of error also causes semantic ambiguity that impedes the understandability of learner texts.

\paragraph{Word Selection} This error occurs when the sentence has wrong words, redundant words or is lack of certain constituents. In the example in Table \ref{tb:AM error types}, the preposition 为 (``for'') in the L2 sentence should be 去 (``to'') which introduces the adjunct of purpose. 
The wrong word selection caused one annotator refuse to label the adjunct behind it.

\paragraph{Ambiguity} This error occurs when some Chinese sentences per se can lead to ambiguity. Sometimes the disagreement of \textit{AM} can be caused by the ambiguity of Chinese itself. 
In the example sentence, ``和妈妈'' can either be ``and mom" in which case it will be part of coordinative agents as \textit{A0} or ``with mom" that serves as an adjunct (\textit{AM}) of the predicate.

\paragraph{Lack of Preposition}
This error occurs when an \textit{AM} requires a preposition while the sentence leaves it out.
The most frequent cause for this error is verb subcategorization, e.g., mistaking interactive verb as non-interactive verb. In the example in Table \ref{tb:AM error types}, the interactive verb 见面 (``meet'') usually has two parties as coordinative subjects (\textit{A0}) linked by an auxiliary word 和 (``with''). 
However, learners often omit 和 (``with'') and put the second \textit{A0} behind the verb which is rather confusing to the annotators. 

\paragraph{\textit{AM-AN} Confounding} This error occurs when the sentence mistakes \textit{AN} as \textit{AM}. In the example, the word 流利 (``fluent'') is a predicate in Chinese whose only argument (\textit{A0}) should be the language that is fluent. 
However, the learner mistook the person who can speak a certain fluent language as \textit{A0} and put the language in a locative \textit{AM}. 
In this case, the annotator cannot decide whether to label the language or the person as \textit{A0}.

\begin{table*}[]
\centering
\caption{Descriptions and percentages of \textit{AM} error types}
\renewcommand{\arraystretch}{1.1}
\scalebox{1.0}{
\begin{tabular}{m{1.5cm}m{7.5cm}m{5.8cm}m{0.6cm}}
\hline
\multicolumn{1}{l}{
Error type} & \multicolumn{1}{c}{
Example} & \multicolumn{1}{c}{
Disagreement} & \multicolumn{1}{c}{
\textbf{\%}} \\\hline
\begin{tabular}[c]{@{}l@{}}Word \\ Order\end{tabular}  & \begin{tabular}[c]{@{}l@{}}

\begin{tabular}{m{0.2cm}m{0.5cm}m{0.8cm}m{1.43cm}l}
\textbf{L2} & 别 & 离开 & 鄂木斯克 & 打算！ \\
 & Don't & leave & Emusike & try! \\
\textbf{L1} & 别 & 打算 & 离开 & 鄂木斯克! \\
 & Don't & try & leave & Emusike!
\end{tabular}

\end{tabular} 
& \begin{tabular}[c]{@{}l@{}}
{[别]}$_{AM}${[离开\ 鄂木斯克]}$_{A1}${[打算]}$_{rel}$！\\ 
{[别\ \ \ \ \ \ \ \ \ 离开\ 鄂木斯克]}$_{A1}${[打算]}$_{rel}$！
\end{tabular} & 39\% \\
\hline
\begin{tabular}[c]{@{}l@{}}Word \\ Selection\end{tabular} & \begin{tabular}[c]{@{}l@{}}

\begin{tabular}{m{0.2cm}m{0.1cm}m{0.3cm}m{0.9cm}m{0.3cm}m{0.3cm}m{0.8cm}l}
\textbf{L2} & 我 & 被 & 召唤 & 为 & 帮 & 他们 & 翻译。 \\
 & I & pass & summon & for & help & they.acc & translate.\\
\textbf{L1} & 我 & 被 & 召唤 & 去 & 帮 & 他们 & 翻译。 \\
 & I & pass & summon & for & help & they.acc & translate.
\end{tabular}\\

\end{tabular} 
& \begin{tabular}[c]{@{}l@{}}
{[我]}$_{A1}$ 被{[召唤]}$_{rel}$ 为帮他们翻译。 \\
{[我]}$_{A1}$ 被{[召唤]}$_{rel}$ {[为帮他们翻译]}$_{AM}$。
\end{tabular} & 27\% \\
\hline
\begin{tabular}[c]{@{}l@{}}Ambiguity \end{tabular} & 
\begin{tabular}[c]{@{}l@{}}

\begin{tabular}{m{0.2cm}m{0.2cm}llll}
\textbf{L2} & 我 & 和 & 妈妈 & 一起 & 住。 \\
 & I & with & mom & together & live.
\end{tabular}

\end{tabular} 
& \begin{tabular}[c]{@{}l@{}}
{[我\ \ \ \ \ \ \ 和\ 妈妈]}$_{A0}$\ \ {[一起]}$_{AM}${[住]}$_{rel}$。\\ 
{[我]}$_{A0}${[和\ 妈妈]}$_{AM}${[一起]}$_{AM}${[住]}$_{rel}$。\\ 
\end{tabular} & 16\% \\
\hline
\begin{tabular}[c]{@{}l@{}}Lack of \\ proposition\end{tabular}  & 
\begin{tabular}[c]{@{}l@{}}

\begin{tabular}{m{0.2cm}m{0.1cm}m{1cm}m{1cm}m{0.8cm}m{1cm}l}
\textbf{L2} & 我 & 昨天 & 见面 & 他们 & 了。 &  \\
 & I & yesterday & meet.past & they.acc & asp. &  \\
\textbf{L1} & 我 & 昨天 & 和 & 他们 & 见面 & 了。 \\
 & I & yesterday & with & they.acc & meet.past & asp.
\end{tabular}

\end{tabular}
& \begin{tabular}[c]{@{}l@{}}
{[我]}$_{A0}${[昨天]}$_{AM}${[见面]}$_{rel}${[他们]}$_{AM}$了。\\ 
{[我]}$_{A0}${[昨天]}$_{AM}${[见面]}$_{rel}$他们\ \ \ \ \ \ \ \ \ 了。\\
\end{tabular} & 10\% \\
\hline
\begin{tabular}[c]{@{}l@{}}\textit{AM-AN} \\ confounding\end{tabular} &  
\begin{tabular}[c]{@{}l@{}}

\begin{tabular}{m{0.2cm}m{0.1cm}m{1.1cm}lll}
\textbf{L2} & 我 & 想 & 流利 & 在 & 日语。 \\
 & I & want to & be fluent & in & Japanese. \\
\textbf{L1} & 我 & 想 & 日语 & \multicolumn{2}{l}{流利。} \\
 & I & want & my Japanese & \multicolumn{2}{l}{to be fluent.}
\end{tabular}

\end{tabular} 
& \begin{tabular}[c]{@{}l@{}}
 {[我]}$_{A0}$想{[流利]}$_{rel}${[在\ \ 日语]}$_{AM}$。\\ 
我\ \ \ \ \ \ \ \ 想{[流利]}$_{rel}$在{[日语]}$_{A0}$。\\
\end{tabular} & 8\% \\
\hline
\end{tabular}}
\label{tb:AM error types}
\end{table*}

\section{Conclusion}
In  this paper, we present an L2-L1 parallel corpus for SRL on learner Chinese texts. This is achievable since the learner Chinese texts are quite understandable. Such a corpus can be applied to analyzing error patterns in terms of argument structure as well as evaluating the performance of NLP systems. 
In our case study on learner Chinese texts by speakers of four typological mother tongues, the errors are mainly caused by the mismatch between the learners' mother language and Chinese grammar, including word order, word selection, lack of proposition and argument-adjunct confounding. 
We also apply this L2-L1 corpus to certain automatic labeling tasks to give a sample evaluation on the potential of such resources.
More detailes can found in the sister paper of this work \citep{Lin:18}.

\end{CJK*}
\bibliographystyle{acl_natbib_nourl}
\bibliography{references}
\end{document}